# RURA-Net: A general disease diagnosis method based on Zero-Shot Learning

Yan Su[1,†], Qiulin Wu[1,†], Weizhen Li[1], Chengchang Pan[1,*], Honggang Qi[1,*]

[1] School of Computer Science and Technology, University of Chinese Academy of Sciences
†Co-first author
*corresponding author (`chpan.infante@qq.com, hgqi@ucas.ac.cn`)

**Abstract.** The training of deep learning models relies on a large amount of labeled data. However, the high cost of medical labeling seriously hinders the development of deep learning in the medical field. Our study proposes a general disease diagnosis approach based on Zero-Shot Learning. The Siamese neural network is used to find similar diseases for the target diseases, and the U-Net segmentation model is used to accurately segment the key lesions of the disease. Finally, based on the ResNet-Agglomerative clustering algorithm, a clustering model is trained on a large number of sample data of similar diseases to obtain a approximate diagnosis of the target disease. Zero-Shot Learning of the target disease is then successfully achieved. To evaluate the validity of the model, we validated our method on a dataset of ophthalmic diseases in CFP modality. The external dataset was used to test its performance, and the accuracy=0.8395, precision=0.8094, recall=0.8463, F1 Score=0.8274, AUC=0.9226, which exceeded the indexes of most Few-Shot Learning and One-Shot Learning models. It proves that our method has great potential and reference value in the medical field, where annotation data is usually scarce and expensive to obtain.

## 1 Introduction

Medical imaging constitutes an essential part of the clinical workflow for understanding and intervening in disease [1], with AI serving as a "second pair of eyes" for doctors to perform imaging measurements, and quickly identify disease patterns and trends through image analysis technology [2]. For example, the accuracy of deep neural networks in a variety of applications has matched or exceeded that of clinical experts [3], as demonstrated in referral recommendations for sight-threatening retinal diseases [4] and pathological detection of chest X-ray images [5]. Nevertheless, effective training of machine learning models, especially deep learning models, depends on access to large and high-quality manual labeling data, which creates a considerable burden and seriously hinders the development of deep learning in the medical field due to the demanding workload and expert evaluation with a high threshold [6-9].

Zero-shot Learning (ZSL) is an innovative concept and learning technique that builds recognition models by transferring knowledge of known categories without the need to take samples from previously unseen categories during training [10-12]. For example, ZSL-based models will be able to automatically learn and diagnose COVID-19 patients based on existing chest X-ray images of patients with asthma and lung inflammatory diseases that clinicians have already identified and labeled, as well as some new images [13]. These models promise to transcend the conventional dependency of supervised learning on extensive data labeling, thereby expediting the discovery of novel disease identification and treatment methodologies.

However, the application of ZSL in the medical field has not been fully studied. Therefore, we propose a new general disease diagnosis framework based on ZSL, named RURA-Net, and systematically evaluated its performance and versatility using CFP modality ophthalmic diseases as an example (see Fig. 1). Specifically, for the target disease, which may be a rare disease, our model RURA-Net will first find the disease with the highest similarity to the target disease through the Siamese twin neural network, and its samples may be relatively easy to obtain. The clinical diagnosis lesions of the target disease are queried, and the segmentation model for the target disease lesion is obtained by training the U-Net neural network. Subsequently, a large number of similar disease data are input and the segmentation results are used to train a ResNet-Agglomerative clustering network that can approximately diagnose the target disease, and finally achieve ZSL for the target disease. Through the above process, we propose a method to reduce the dependence of clinical diagnosis on data, and also provide a valuable reference for future research improvement in ZSL.

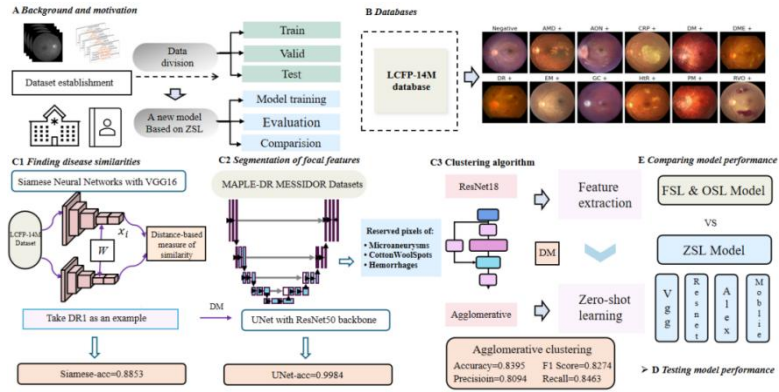

**Fig. 1.** Schematic of development and evaluation of the model based on ZSL.

## 2      Methodology

In summary, our model architecture consists of a Siamese neural network for finding the similarity between diseases, a U-Net neural network for segmenting disease lesions, and a ResNet-Agglomerative fusion network for clustering. Each of these sections will be elaborated in detail in the following.

### 2.1 Siamese Network for Finding Commonalities

The Siamese neural network outputs a score to judge the similarity between the two inputs by feeding them into two similar subnetworks with the same architecture, parameters, and weights. Figure 2 illustrates the architecture of a Siamese network.

The pre-trained VGG19 network within our Siamese architecture extracted deep features from images, and converted them into feature maps. These maps were then compared using the L1 norm to assess similarity. Then, the feature vector is processed through a fully connected layer, followed by a sigmoid activation function to produce a probability value. In the training process, we used Contrastive Loss to optimize the weights, whose expression is as follows:

$$L(W,(Y,X_1,X_2)) = \frac{1}{2N}\sum_{n=1}^{N} Y D_W^2 + (1-Y)\max(m - D_W, 0)^2 \qquad (1)$$

where $D_W(X_1, X_2) = \|X_1 - X_2\|_2 = (\sum_{i=1}^{P}(X_1^i - X_2^i)^2)^{\frac{1}{2}}$ represents the L2 norm of the features $X_1$ and $X_2$, P is the feature dimension of the samples, Y is the label of whether the two samples match, m is the set threshold, and N is the number of samples.

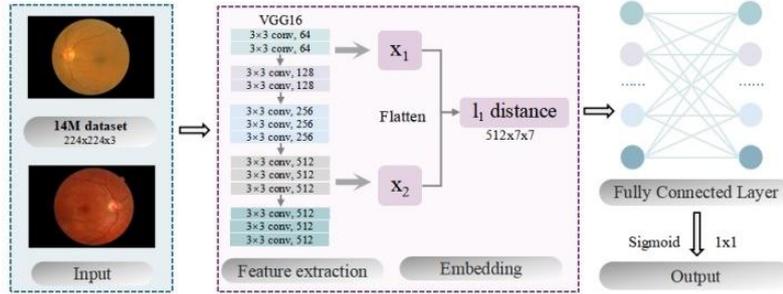

**Fig. 2.** Siamese Network Architecture Diagram

### 2.2 Segmentation Based on U-Net

We implemented U-Net as a specialized feature extracter to accurately segment clinical lesions in retinal images, thereby optimizing the subsequent disease diagnosis task. The architecture is shown in Figure 3.

In our experiment, the feature extractor employed a pre-trained ResNet50 as the convolutional neural network backbone, transforming the input image x of size $512 \times 512 \times 3$ into a feature map $F_{ResNet50}(x)$ of size $16 \times 16 \times 2048$. Feature fusion and processing involved the upsampling module U that upscaled and concatenated the feature map $F_{ResNet50}(x)$ to produce $F_{up}$. The final segmentation result was obtained through a convolutional layer. The training strategy incorporated focal loss to address the imbalance between positive and negative samples.

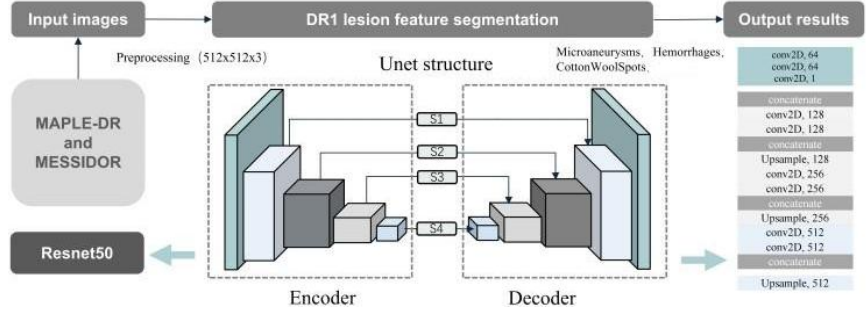

**Fig. 3.** U-Net Architecture for Lesion Feature Segmentation

### 2.3   ResNet-Agglomerative Clustering for ZSL

We designed ResNet-Agglomerative clustering, enabling ZSL by clustering feature vectors without relying on labeled data.

The core of ResNet network is that it proposes the framework of residual learning and introduces a new network structure, namely residual block. Its Skip Connection can effectively alleviate the network degradation problem caused by the deep learning network deepening process. Agglomerative clustering (AGG) is a bottom-up clustering algorithm. The distance measure between data points can be expressed as

$$D = \sqrt{(x_1 - y_1)^2 + (x_2 - y_2)^2} \qquad (2)$$

In our experiment, a pre-trained ResNet18 served as the backbone, and input images were resized to 224×224 pixels and normalized. The final classification layer of ResNet18 was removed to output high-dimensional feature vectors. These feature vectors were then clustered and analyzed using the AGG algorithm.

## 3      Experiments

### 3.1    Datasets

In our experiments, we evaluated the feasibility of the RURA-Net framework on datasets of ophthalmic diseases, all of which are publicly available (see Table 1).

**Table 1.** Details of the datasets used in the experiments

| Dataset | Number | Purpose | Reference |
|---|---|---|---|
| LCFP-14M | 13,718,610 | Find similarity and train clustering network | Qiulin, W., et al. [14] |
| MESSIDOR | 1,200 | Train segmentation network | Decencière, E., et al. [15] |
| MAPLES-DR | 198 | Train segmentation network | Lepetit-Aimon, G., et al. [16] |
| Eyepacs | 35,126 | External validation | Kaggle and EyePacs. [17] |

### 3.2 Evaluation Metrics

The evaluation metrics for our experiment included accuracy, precision, recall, F1 Score, AUC value, mean intersection over union (mIoU), mean pixel accuracy (mPA), and overall accuracy.

### 3.3 Implementation Environment

For our ZSL task, the experiments were simulated using Visual Studio Code as the compiler, Python version 3.9 for programming and PyTorch 11.0.0+cuda11.3 as the development foundation. Training on GPUs including NVIDIA GeForce RTX 3080s, paired with a powerful Intel Xeon E5-2678 v3 CPU.

## 4 Experimental Results

We used Siamese networks to evaluate the similarity between 11 ophthalmic diseases in LCFP-14M dataset and obtain a similarity matrix. Taking mild diabetic retinopathy (DR1) as an example, Degenerative Myopia (DM) was identified as the most relevant disease, a finding that is also consistent with clinical observations that DM is a major risk factor for DR1. The experimental results show that the accuracy of Siamese network is 0.8853, the recall is 0.9012, and the F1 score is 0.8873. These metrics demonstrate the strong performance of the network in judging disease associations.

Based on clinical diagnostic standards, we identified three main features of DR1: Microaneurysms [18], Hemorrhages [19], and Cotton Wool Spots [20]. We trained a segmentation model on a combined dataset of MAPLE-DR and MESSIDOR. For Microaneurysms, the model achieved a mIoU of 56.41%, a mPA of 60.79%, and overall accuracy of 99.84%. Hemorrhages detection showed the highest performance among the features, with a mIoU of 60.17%, a mPA of 63.74%, and an impressive accuracy of 99.97%. For Cotton Wool Spots, the model attained a mIoU of 56.72%, a mPA of 62.15%, and an accuracy of 99.83%. These results indicate that our model can identify and localize key pathological features of DR1 with high precision, especially in terms of overall accuracy (all above 99.8%).

The evaluation index of the clustering model shows that the accuracy of the model reaches 0.8395, and the precision is 0.8094, the recall is 0.8463, the F1 Score is 0.8274, and the AUC reaches 0.9226, which reflects the good overall performance of the model in terms of clustering performance.

### 4.1 Comparison with FSL and OSL

In order to verify the performance of our ZSL model, we compared it with Few-Shot Learning (FSL) model and One-Shot Learning (OSL) model using different backbone networks. By comparing the performance of our ZSL model with FSL and OSL models (see Table 2), we find that our ZSL model is better than most FSL and OSL models in terms of overall average performance, and the overall accuracy is at a good level. This shows the great potential of ZSL in assisting medical diagnosis and alleviating the problem of lack of medical labels. However, it is worth noting that some supervised models such as ResNet18 and ResNet34 perform poorly and end up close to random classification. Some models, such as GoogLeNet and ResNet18, exhibit extremely unbalanced metrics, which indicates that there is a non-negligible risk of underfitting on small sample datasets.

**Table 2.** FSL and ZSL model performance comparison

| Training Set | | Eval Set | | Model | Metrics | | | | |
|---|---|---|---|---|---|---|---|---|---|
| Dataset | #Samples | Dataset | #Samples | | Accuracy | Precision | Recall | F1 Score | AUC |
| LCFP-14M | 56×2 | Eyepacs | 1000×2 | ResNet18 | 0.5235 | 0.5223 | 0.5510 | 0.5363 | 0.5355 |
| | | | | ResNet34 | 0.4855 | 0.4896 | 0.6850 | 0.5711 | 0.5058 |
| | | | | ResNet50 | 0.6485 | 0.6174 | 0.7810 | 0.6896 | 0.7061 |
| | | | | GoogLeNet | 0.5220 | 0.5114 | **0.9900** | 0.6744 | 0.7069 |
| | | | | MobileNet | 0.7340 | 0.7246 | 0.7550 | 0.7395 | 0.8107 |
| | | | | AlexNet | 0.7355 | 0.6630 | 0.9580 | 0.7836 | 0.8681 |
| | | | | VGG16 | **0.8635** | **0.8539** | 0.8770 | **0.8874** | **0.9354** |
| | | | | ZSL | **0.8395** | **0.8094** | 0.8463 | 0.8274 | 0.9226 |

### 4.2 Backbone Network Evaluation

When replacing the backbone network of Siamese network, the performance of ResNet series and VGG series improves with the increase of architecture complexity, among which VGG19 performs significantly best with an accuracy of 0.8853, a recall of 0.9012, and a F1 score of 0.8873. However, the EfficientNet and DenseNet series both have overfitting problems to varying degrees. The results of the experiment are shown in Table 3.

**Table 3.** Evaluating the performance of various backbone models for disease relevance assessment using Siamese networks

| Training Set | | Eval Set | | Backbone | Metrics | | |
|---|---|---|---|---|---|---|---|
| Dataset | #Samples | Dataset | #Samples | | Accuracy | Recall | F1 Score |
| 14M | 3577 | Eyepacs | 5000 | ResNet18 | 0.1068 | 0.1068 | 0.1931 |
| | | | | ResNet34 | 0.4918 | 0.4918 | 0.6593 |
| | | | | ResNet50 | 0.4578 | 0.4578 | 0.6281 |
| | | | | ResNet101 | 0.5642 | 0.5642 | 0.7214 |
| | | | | ResNet152 | 0.8691 | 0.8691 | 0.9300 |
| | | | | VGG16 | 0.5442 | 0.5442 | 0.7048 |
| | | | | **VGG19** | **0.8853** | **0.9012** | **0.8873** |
| | | | | MobileNet-V2 | 0.7443 | 0.7443 | 0.8534 |
| | | | | MobileNet-V3 | 0.8940 | 0.8940 | 0.9440 |
| | | | | EfficientNet-B0 | 0.7707 | 0.7707 | 0.8705 |
| | | | | EfficientNet-B4 | overfitting | overfitting | overfitting |
| | | | | DenseNet121 | 0.2589 | 0.2589 | 0.4113 |
| | | | | DenseNet169 | 0.8407 | 0.8407 | 0.9135 |
| | | | | DenseNet201 | overfitting | overfitting | overfitting |

RURA-Net: A general disease diagnosis method based on Zero-Shot LearningWhen replacing the backbone network of the clustering algorithm, ResNet18 has higher indicators, showing a relatively balanced performance. Its accuracy is 0.8395, precision is 0.8094, recall is 0.8463, and F1 Score is 0.8274. It is worth noting that ResNeXt has a high recall rate of 0.9248, but an accuracy rate of only 0.6839 and an F1 Score of 0.7863, indicating that this backbone network is not suitable for the algorithm of this experiment. The results of the experiment are shown in Table 4.

**Table 4.** Comparison of different backbone networks for clustering algorithms

| Training Set | | Eval Set | | Model | Metrics | | | |
|---|---|---|---|---|---|---|---|---|
| Dataset | #Samples | Dataset | #Samples | | Accuracy | Precision | Recall | F1 Score |
| LCFP-14M | 3577 | Eyepacs | 5000 | **ResNet18** | **0.8395** | **0.8094** | **0.8463** | **0.8274** |
| | | | | ResNet34 | 0.8307 | 0.7931 | 0.8490 | 0.8201 |
| | | | | ResNet50 | 0.8307 | 0.7931 | 0.8490 | 0.8201 |
| | | | | ResNet101 | 0.8371 | 0.8018 | 0.8523 | 0.8263 |
| | | | | ResNet152 | 0.8078 | 0.7307 | 0.9141 | 0.8122 |
| | | | | GoogLeNet | 0.8148 | 0.7577 | 0.8711 | 0.8105 |
| | | | | AlexNet | 0.7926 | 0.7593 | 0.7960 | 0.7772 |
| | | | | MobileNet | 0.8252 | 0.8602 | 0.7349 | 0.7926 |
| | | | | ResNeXt | 0.7715 | 0.6839 | 0.9248 | 0.7863 |
| | | | | ViT | 0.6855 | 0.8964 | 0.3483 | 0.5017 |

### 4.3 Clustering Algorithm Comparison

In this experiment, the performances of several different clustering algorithms were compared, including KMeans, Gaussian Mixture Model (GMM), AGG, CLARA, KModes, Partitioning Around Medoids (PAM), and KMedoids.

As shown in the Table 5, the AGG algorithm outperformed the other algorithms, maintaining high metrics across the board. Its accuracy, precision, recall, and F1 Score were 0.8395, 0.8094, 0.8463, and 0.8274 respectively. This suggested that the model could achieve high-precision clustering results and was suitable for our model. However, KModes had a higher degree of misjudgment. And the poor performance of GMM may be due to the fact that GMM assumes a Gaussian distribution, which is not consistent with the actual distribution of the current dataset.

**Table 5.** Comparison of different clustering algorithms in the classification of eye diseases.

| Training Set | | Eval Set | | Model | Metrics | | | |
|---|---|---|---|---|---|---|---|---|
| Dataset | #Samples | Dataset | #Samples | | Accuracy | Precision | Recall | F1 Score |
| LCFP-14M | 3577 | Eyepacs | 5000 | KMeans | 0.8337 | 0.8700 | 0.7456 | 0.8030 |
| | | | | GMM | 0.5491 | 1.0000 | 0.0081 | 0.0160 |
| | | | | **AGG** | **0.8395** | **0.8094** | **0.8463** | **0.8274** |
| | | | | CLARA | 0.8469 | 0.8601 | 0.7919 | 0.8246 |
| | | | | KModes | 0.7514 | 0.6545 | 0.9597 | 0.7782 |
| | | | | PAM | 0.8414 | 0.8166 | 0.8396 | 0.8279 |
| | | | | KMedoid | 0.8380 | 0.8072 | 0.8456 | 0.8260 |

### 4.4    Ablation Study

To better understand the contribution and necessity of disease similarity analysis and lesion segmentation to the ZSL model's superior performance, we performed two ablation experiments. In the first phase, without focusing on disease similarities. The lower performance indicated that the clustering model's effectiveness decreased without similarity analysis. In the second phase, we directly extracted 5,000 images with DM=1 from the dataset, without using U-Net for segmentation. The model's accuracy and recall rate both significantly dropped to 0.3074. This suggested that removing the U-Net lesion segmentation step caused the model to misclassify most negative samples, leading to strong bias. Therefore, both the disease similarity analysis and lesion segmentation steps were crucial for achieving robust performance of the ZSL model. The results of the experiment are shown in Table 6.

**Table 6.** Ablation Experiment Results for ZSL Model

| Method | Accuracy | Precision | Recall | F1 Score | ROC Area |
|---|---|---|---|---|---|
| Segmentation | 0.6169 | 0.6342 | 0.6169 | 0.6041 | 0.6169 |
| Siamese | 0.3074 | 0.9163 | 0.3074 | 0.3908 | 0.5974 |
| Siamese+Segmentation | 0.8337 | 0.8700 | 0.7456 | 0.8030 | 0.9226 |

## 5    Conclusion

This study introduces RURA-Net, a new ZSL-based method for general disease diagnosis and demonstrates its effectiveness in identifying various diseases without relying on a large labeled data set by using ophthalmic diseases as validation. RURA-Net starts with a Siamese neural network, which aims to reveal the correlation between different diseases. The pre-trained U-Net segmentation model with ResNet50 as the backbone is used next for accurate segmentation of key pathological lesion. The final step of RURA-Net is using the ResNet-Agglomerative clustering algorithm to classify the disease states in an unsupervised manner. The comparative and ablation experiments show that our ZSL model achieves competitive results. This highlights the potential of ZSL as an alternative in situations where labeled data is scarce or unavailable, especially compared to traditional supervised learning models.

However, our model still has certain limitations. It is currently only applicable to single-modality image data, and has the common limitation of many deep learning-based diagnostic models, which is poor interpretability. To advance this field in the future, we can start with developing ZSL models that can effectively handle multiple imaging modalities, and strive to improve the interpretability of the models. In general, our method can be transferred to new, annotated data, showing great potential in the medical field where annotated data is usually scarce and expensive to obtain, providing extraordinary reference value for subsequent research on the application of artificial intelligence in the medical field.